\documentclass[runningheads]{llncs}
\usepackage[T1]{fontenc}
\usepackage{graphicx}
\usepackage{algorithm}
\usepackage{algorithmic}
\usepackage{amsmath}
\usepackage{amsfonts}
\usepackage{amssymb}
\usepackage{bbm}
\usepackage{booktabs}
\usepackage{multirow}

\makeatletter
\newcommand{\figcaption}[1]{\def\@captype{figure}\caption{#1}}
\newcommand{\tblcaption}[1]{\def\@captype{table}\caption{#1}}
\makeatother

\newcommand{\ngrad}{{\tilde\nabla}}
\newcommand{\argmax}{\operatornamewithlimits{argmax}} 
\newcommand{\diag}{\operatornamewithlimits{diag}} 
\newcommand{\E}{\mathbb{E}} 
\newcommand{\D}{\mathcal{D}} 
\newcommand{\Loss}{\mathcal{L}} 
\newcommand{\F}{\mathcal{F}}
\newcommand{\G}{\mathcal{G}} 
\newcommand{\R}{\mathcal{R}} 

\newcommand{\T}{\mathrm{T}} 
\newcommand{\I}{\mathbb{I}} 
\newcommand{\Q}{\mathcal{Q}} 

\begin{document}

\title{Efficient Search of Multiple Neural Architectures with Different Complexities via Importance Sampling}
\titlerunning{Efficient Search of Multiple Neural Architectures with Different Complexities}
%
\author{Yuhei Noda\inst{1} \orcidID{0000-0001-9646-2506}
\and
Shota Saito\inst{1,2} \orcidID{0000-0002-9863-6765}
\and
Shinichi Shirakawa\inst{1} \orcidID{0000-0002-4659-6108}
}
\authorrunning{Y. Noda et al.}
%
\institute{Yokohama National University, Kanagawa, Japan\\ \email{nodayuhei@gmail.com, saito-shota-bt@ynu.jp, shirakawa-shinichi-bg@ynu.ac.jp}
\and SkillUp AI Co., Ltd., Tokyo, Japan\\ \email{s\_saito@skillupai.com}
}
\maketitle
\begin{abstract}
Neural architecture search (NAS) aims to automate architecture design processes and improve the performance of deep neural networks. Platform-aware NAS methods consider both performance and complexity and can find well-performing architectures with low computational resources. Although ordinary NAS methods result in tremendous computational costs owing to the repetition of model training, one-shot NAS, which trains the weights of a supernetwork containing all candidate architectures only once during the search process, has been reported to result in a lower search cost. This study focuses on the architecture complexity-aware one-shot NAS that optimizes the objective function composed of the weighted sum of two metrics, such as the predictive performance and number of parameters. In existing methods, the architecture search process must be run multiple times with different coefficients of the weighted sum to obtain multiple architectures with different complexities. This study aims at reducing the search cost associated with finding multiple architectures. The proposed method uses multiple distributions to generate architectures with different complexities and updates each distribution using the samples obtained from multiple distributions based on importance sampling. The proposed method allows us to obtain multiple architectures with different complexities in a single architecture search, resulting in reducing the search cost. The proposed method is applied to the architecture search of convolutional neural networks on the CIAFR-10 and ImageNet datasets. Consequently, compared with baseline methods, the proposed method finds multiple architectures with varying complexities while requiring less computational effort.
\keywords{Neural Architecture Search  \and Convolutional Neural Network \and Importance Sampling \and Natural Gradient}
\end{abstract}

\section{Introduction}
Architecture design is a key factor in accelerating the performance of deep neural networks (DNNs); however, the associated process is arduous for practitioners. Neural architecture search (NAS), aimed at automating the design of DNN architectures, has been actively studied in recent years~\cite{Elsken2018}. Popular methods often optimize architectures using evolutionary algorithms~\cite{Real2017,Suganuma2017} or reinforcement learning~\cite{Zoph2017}. These early NAS methods optimize the architecture in hyperparameter optimization frameworks, which requires a significant amount of time for architecture search due to the repetition of model training. \textit{One-shot NAS}, e.g., \cite{Liu2019,Pham2018,Shirakawa2018}, is a promising approach for reducing the computational cost of NAS. One-Shot NAS simultaneously optimizes the weight and architecture parameters considering an extensive network (supernetwork) that includes many candidate architectures as its subnetworks. Because the weight parameters are shared between subnetworks, one-shot NAS optimizes the weights in the supernetwork only once during the search process, thus significantly reducing the search cost.

DNNs are often implemented in devices with limited computational resources, such as embedded and mobile devices. In such cases, NAS methods are required to find an architecture with good prediction performance and low computation and memory usage. Therefore, NAS methods have been developed for optimizing the prediction performance and architecture complexity, such as FLOPs, latency, and the number of weight parameters. The method proposed in \cite{wu2018fbnet} includes a term related to the latency in the objective function and successfully finds a highly accurate architecture while suppressing latency. This study focuses on the one-shot NAS method proposed in \cite{Saito2019}, which introduces a regularization of the architecture complexity. This method uses binary variables to represent the architecture and a multivariate Bernoulli distribution as the law of binary variables. The architecture search is performed by updating the parameters of the Bernoulli distribution to minimize the weighted sum of the predictive loss and the regularization term related to the complexity of the architecture. Although this approach reduces the number of connections in a densely connected convolutional neural network (CNN), the architecture search space is limited because it must represent the architecture using binary variables. This implies that directly employing the state-of-the-art architecture search space represented by categorical variables~\cite{Pham2018,Zoph2018} is not straightforward. Additionally, obtaining multiple architectures with different complexities requires repeated architecture searches with different regularization coefficients in the objective function, which, in turn, increases computational cost.

This study extends the method proposed in \cite{Saito2019} to overcome the limitation described above. We adopt categorical distributions as the architecture distribution and propose a regularization term for the architecture complexity. We also derive the analytical natural gradient of the proposed regularization term. Thereafter, we propose an efficient search method to simultaneously obtain multiple architectures with different complexities in a single architecture search using importance sampling. The proposed method is then evaluated by applying it to the architecture search of CNNs on the CIFAR-10~\cite{Krizhevsky09learningmultiple} and ImageNet~\cite{deng2009imagenet} datasets. The experimental results indicate that the proposed method can obtain multiple architectures with different complexities in a single search, and its prediction performance is comparable with that demonstrated by baseline methods.

\section{Probabilistic Model-Based One-Shot NAS with Complexity Regularization}
This section details the one-shot NAS framework using the architecture complexity regularization proposed in \cite{Saito2019}. We denote the DNN parameterized by architecture parameters $M$ and weights $W$ as $\phi(M, W)$ and assume that $\phi$ is differentiable with respect to (w.r.t.) $W$ but non-differentiable w.r.t. $M$. The architecture parameters $M$ determine the types of connections and operations in DNN. The architecture defined by $M$ corresponds to a subnetwork in the supernetwork and shares weights in the operations between different architectures.

Let us consider the optimization of $W$ and $M$ to minimize both the loss, for instance, the cross-entropy loss, and the regularization metric w.r.t. the complexity of the architecture. We denote the loss for the dataset $\D$ and the regularization term as $\Loss(M, W, \D)$ and $\R(M)$, respectively. In \cite{Saito2019}, the weighted sum of the two terms, $\F(M, W) = \Loss(M, W, \D) + \epsilon \R(M)$, has been adopted as the objective function, where $\epsilon$ represents the regularization coefficient that balances the two terms. However, because the architecture parameters $M$ are non-differentiable and often discrete, we cannot optimize $M$ by a gradient method. To relax the problem, we introduce the parametric probability distribution of $M$ and denote it as $P_{\theta}(M)$, where $\theta$ denotes the distribution parameters. Instead of directly optimizing $\F(M, W)$, we optimize $\theta$ by minimizing the expected loss of $\F(M, W)$ under $P_{\theta}(M)$ as $\G (\theta ,W)=\E_{{P}_\theta}\left [ \Loss (M,W,\D) \right ]+\epsilon \E_{{P}_\theta }\left [\R (M)\right]$.

As the objective function $\G (\theta, W)$ is differentiable w.r.t. both $W$ and $\theta$, we can optimize it by a gradient method. We update the distribution parameters $\theta$ to the natural gradient direction~\cite{Amari1998}, which is the steepest direction when considering the Kullback–Leibler divergence as the pseudo-distance in the distribution parameter space, and it is given by the product of the inverse of the Fisher information matrix (FIM) and Euclidean gradient. We use the vanilla gradient to optimize $W$ as usual DNN training. The gradients w.r.t. $W$ and $\theta$ are given by
\begin{align}
& \nabla_{W} \G(\theta,W) = \E_{{P}_{\theta}} \left[\nabla_W\Loss(M,W,\D) \right] \label{eq:omomi_gradient}\\
& \ngrad_{\theta} \G(\theta,W) = \E_{{P}_{\theta}} \left[\Loss(M,W,\D)\ngrad_{\theta}\ln P_{\theta}(M) \right]+\epsilon \ngrad_{\theta} \E_{{P}_\theta} \left[\R (M)  \right] \enspace , \label{eq:theta_natural_gradient}
\end{align}
where $\ngrad_{\theta}=F(\theta)^{-1}\nabla_{\theta}$ represents the natural gradient operator. Here, $F(\theta)$ indicates the FIM of $P_{\theta}(M)$. Optimizing $\theta$ using \eqref{eq:theta_natural_gradient} with $\epsilon = 0$ operates in a manner similar to information geometric optimization \cite{Ollivier2017}, which is a unified framework for probabilistic model-based evolutionary algorithms. In most cases, it is difficult to compute the gradients \eqref{eq:omomi_gradient} and \eqref{eq:theta_natural_gradient}. Therefore, the gradients \eqref{eq:omomi_gradient} and \eqref{eq:theta_natural_gradient} are approximated using Monte Carlo methods with $\lambda$ architecture parameters $M_1, M_2, \dots, M_\lambda$ sampled from $P_{\theta}(M)$ as follows:
\begin{align}
\nabla_{W} \G(\theta,W) \approx &\frac{1}{\lambda}\sum_{i=1}^{\lambda}\nabla_W\Loss(M_i,W,\D)\label{eq:omomi_gradient2}\\
\ngrad_{\theta} \G(\theta,W)\approx&\frac{1}{\lambda}\sum_{i=1}^{\lambda}\Loss(M_i,W,\D)
\ngrad_{\theta}\ln p_{\theta}(M_i)+\epsilon\ngrad_{\theta} \E_{{P}_\theta} \left[\R (M) \right] \enspace .
\end{align}

Because the scale of the loss affects the magnitude of the natural gradient, we transform $\Loss(M_i,W, \D)$ into the quantile-based utility value under $P_{\theta}(M)$, as done in \cite{Ollivier2017}. The probability of sampling a solution with a loss value less than or equal to $\Loss(M_i,W, \D)$ is estimated as $\bar{q}_{\theta}^{\leqslant}(M_i) =\lambda^{-1}\sum_{k=1}^{\lambda} \I\{\Loss (M_k,W,\D) \leqslant \Loss(M_i,W, \D)\}$, where $\I\{\cdot\}$ denotes the indicator function. We use the utility function of $\hat{s}_i=w(\bar{q}_{\theta}^{\leqslant}(M_i))$, instead of $\Loss (M_i,W,\D)$, to update the distribution parameters $\theta$.\footnote{This utility definition does not assume the possibility of sampling architectures with the same loss value. Although it could happen in our case, we use this utility definition for simplicity. A rigorous definition can be found in \cite{Ollivier2017,shirakawa2018sample}.} Specifically, we use the following function for $w$.
\begin{align*}
    w(x) = \begin{cases}
        -2 & (x \leqslant 0.25) \\
        0 & (0.25 < x \leqslant 0.75) \\
        2 & (0.75 < x ) 
    \end{cases}
\end{align*}
Consequently, the update rule for $\theta$ at the $t$-th iteration is given by
\begin{equation}
\theta^{(t+1)}=\theta^{(t)}-\eta\left(\frac{1}{\lambda}\sum_{i=1}^{\lambda}\hat{s}_i
\ngrad_{\theta}\ln p_{\theta}(M_i)+\epsilon\ngrad_{\theta} \E_{{P}_\theta} \left[\R (M) \right]\right) \enspace , \label{eq:theta_natural_gradient2}
\end{equation}
where $\eta$ represents the learning rate for $\theta$. We note that the weights $W$ can be updated using any stochastic gradient descent (SGD) method with \eqref{eq:omomi_gradient2}.

\section{Proposed Method}
In \cite{Saito2019}, the binary vector has been adopted as the architecture parameter. However, state-of-the-art architecture search spaces, such as \cite{Liu2019,Pham2018}, are defined using categorical variables. In addition, repeating the architecture search is required to obtain multiple architectures with different complexities. We first introduce the categorical distribution as $P_{\theta}(M)$ in the framework considered in  \cite{Saito2019}. Subsequently, we propose simultaneously optimizing multiple categorical distributions, each corresponding to a different regularization coefficient, to obtain multiple architectures with varying complexities in a single search. Each categorical distribution is updated by exploiting samples from other distributions to realize an efficient search process.

\subsection{Introducing Categorical Distributions}
The DNN architecture is represented by the following $D$ dimensional categorical variables: $h = (h_1, \dots, h_D)$. The $d$-th categorical variable $h_d$ possesses $K_d$ candidate categories and determines operations or connections in the DNN. For instance, one can determine the kernel size of a convolution layer. We denote categorical variables by one-hot vectors as $M = (m_1, \dots, m_D)$, where $m_d = (m_{d,1}, \dots, m_{d,K_d})^{\T} \in \{0,1\}^{K_d}$. When $h_d$ is the $k$-th category, $m_{d,k} = 1$, and other elements of $m_d$ are zero. We consider the categorical distribution as the distribution of the architecture parameters, which is described as $P_{\theta}(M)=\prod_{d=1}^{D}\prod_{k=1}^{K_d}\left(\theta_{d,k}\right)^{m_{d,k}}$, where $\theta_{d, k} \in [0, 1]$ is the probability of being $m_{d,k} = 1$.

We choose the number of weight parameters as the complexity metric for the regularization term $\R(M)$ to penalize the complicated architecture. Let us denote the number of weight parameters in the operation corresponding to $m_{d,k}$ as $c_{d,k}$; then, we define the regularization term as $\R(M) =\sum_{d=1}^{D}\sum_{k=1}^{K_d}c_{d,k}m_{d,k}$. The expected value of $\R(M)$ under $P_{\theta}(M)$ is described as
\begin{equation}
\E_{P_{\theta}}\left[\R(M)\right]=\sum_{d=1}^{D}\sum_{k=1}^{K_d}c_{d,k}\theta_{d,k} \enspace . \label{eq:rm_ex}
\end{equation}
It should be noted that the distribution parameter of the last category can be given by $\theta_{d, K_d} = 1 - \sum_{k=1}^{K_d-1} \theta_{d, k}$ owing to $\sum_{k=1}^{K_d} \theta_{d, k} = 1$; consequently, we can introduce the notation of the distribution parameter vector without the last category's parameter as $\bar{\theta}_d=(\theta_{d,1},\theta_{d,2},\dots,\theta_{d,K_d-1})^{\T}$.

Next, we derive the natural gradient of $\E_{P_{\theta}}\left[\R(M)\right]$. The vanilla gradient of $\E_{P_{\theta}}\left[\R(M)\right]$ w.r.t the $d$-th distribution parameters $\bar{\theta}_d$ is given by $\nabla_{\bar{\theta}_d}\E_{P_{\theta}}\left[\R(M)\right]=\bar{c}_d-c_{d,K_d}\mathbf{1}$, where $\bar{c}_d=(c_{d,1},c_{d,2},\dots,c_{d,K_d-1})^{\T}$, and $\mathbf{1}$ represents the all-ones vector. The FIM is a block diagonal matrix because our categorical variables are independent. The inverse of the $d$-th block in the FIM is given by $F(\bar{\theta}_d)^{-1} = \diag(\bar{\theta}_d) - \bar{\theta}_{d} \bar{\theta}_{d}^{\T}$. Then, we can obtain the natural gradient of \eqref{eq:rm_ex} as $\ngrad_{\bar{\theta}_d} \E_{P_\theta} \left[ \mathcal{R}(M) \right] = \bar{c}_d \odot \bar{\theta}_d - \left(\bar{c}_d^\T \bar{\theta}_d + c_{d, K_d} (1 - \bar{\theta}_d^\T {\bf 1}) \right) \bar{\theta}_d = \left(\bar{c}_d - \mathcal{Q}_d {\bf 1}  \right) \odot \bar{\theta}_d$, where $\odot$ indicates the element-wise product and $\mathcal{Q}_d = \sum_{k=1}^{K_d} c_{d, k} \theta_{d, k}$. According to \cite{AkimotoICML2019}, the natural gradient of the log-likelihood is given by $\ngrad_{\bar{\theta}_d} \ln P_{\theta}(M) = \bar{m}_d - \bar{\theta}_d$, where $\bar{m}_{d}=(m_{d,1},\dots,m_{d,K_d-1})^{\T}$. We then obtain the update rule of $\bar{\theta}_d$ as
\begin{equation}
\bar{\theta}_{d}^{(t+1)}=\bar{\theta}_{d}^{(t)}-\eta\left(\frac{1}{\lambda}\sum_{i=1}^{\lambda}\hat{s}_i(\bar{m}_{d}^{(i)}-\bar{\theta}_{d}^{(t)})+\epsilon \left(\bar{c}_{d} - \mathcal{Q}_{d} {\bf 1}  \right) \odot \bar{\theta}_{d}^{(t)}\right) \enspace , \label{eq:theta_kousinsiki}
\end{equation}
where $\bar{m}_{d}^{(i)}$ indicates the $d$-th one-hot vector without the last element of the $i$-th sample. Additionally, $\theta_{d,K_d}^{(t+1)} = 1-\sum_{k=1}^{K_d-1}\theta_{d,k}^{(t+1)}$ is given by 
\begin{align}
\theta_{d,K_d}^{(t+1)} 
 =\theta_{d,K_d}^{(t)}-\eta \left(\frac{1}{\lambda}\sum_{i=1}^{\lambda}\hat{s}_{i}\left(m_{d,K_d}^{(i)}-\theta_{d,K_d}^{(t)} \right)+\epsilon\left(c_{d,K_d}-\Q_d \right)\theta_{d,K_d}^{(t)} \right) \enspace . \label{eq:theta_update_last}
\end{align}
According to \eqref{eq:theta_kousinsiki} and \eqref{eq:theta_update_last}, we can replace $\bar{\theta}_d$ in \eqref{eq:theta_kousinsiki} with $\theta_d=(\theta_{d,1},\theta_{d,2},\dots,\theta_{d,K_d})^{\T}$ and update the distribution parameter $\theta_d$ using the replaced update rule.

\subsection{Searching Multiple Architectures via Importance Sampling}
In existing methods \cite{Cai2019,Saito2019}, a search for the architecture must be performed multiple times by altering the regularization coefficient to obtain multiple architectures with different complexities. Herein, we propose a method for finding multiple architectures within a single search, thereby reducing the search cost. The idea is to jointly update the multiple distributions corresponding to different complexities by exploiting the samples drawn from other distributions via importance sampling. Let us consider $N$ distribution parameters, $\boldsymbol{\theta} = (\theta^{(1)},\dots,\theta^{(N)})$, corresponding to different regularization coefficients $\epsilon_1,\dots,\epsilon_N$. The objective function of each distribution is defined by $\G (\theta^{(n)} ,W)=\E_{{P}_{\theta^{(n)}}}\left [ \Loss (M, W, \D) \right ] + \epsilon_n \E_{{P}_{\theta^{(n)}} }\left [\R (M)\right]$. We sample $\lambda$ architecture parameters from the mixture distribution $P_{\boldsymbol{\theta}}(M)=N^{-1}\sum_{n=1}^{N}P_{\theta^{(n)}}(M)$ at each iteration and update each distribution using the samples obtained from the mixture. Based on the importance sampling technique used in \cite{Shirakawa2015gecco,shirakawa2018sample}, the probability $\bar{q}_{\theta^{(n)}}^{\leqslant}(M_i)$ is estimated by
\begin{align}
\bar{q}_{\theta^{(n)}}^{\leqslant}(M_i) = \frac{1}{\lambda} \sum_{k=1}^{\lambda} r_k^{(n)} \I \left\{ \Loss\left(M_k,W,\D\right) \leqslant \Loss\left(M_i,W,\D\right) \right\} \enspace ,
\end{align}
where $r_k^{(n)} = \frac{P_{\theta^{(n)}}(M_k)}{P_{\boldsymbol{\theta}}(M_k)}$ indicates the likelihood ratio.
Then, the utility of $\hat{s}_i^{(n)}= w(\bar{q}_{\theta^{(n)}}^{\leqslant}(M_i))$ is used to update $\theta^{(n)}$. Similarly, the natural gradient can be approximated via importance sampling, and we obtain the update rule of $\theta_d^{(n)}$ as
\begin{align}
\theta_{d}^{(n)} \gets \theta_{d}^{(n)}-\eta\left(\frac{1}{\lambda}\sum_{i=1}^{\lambda} \hat{s}_{i}^{(n)} r_i^{(n)} \left( m_{d}^{(i)} - \theta_d^{(n)} \right)
+\epsilon_n \left(c_d - \mathcal{Q}_d {\bf 1}  \right) \odot \theta_d^{(n)} \right) \enspace . \label{eq:is_theta_kousin}
\end{align}
Here, we ignore the notation of the time step $t$ for simplicity.

\begin{algorithm}[t]
\caption{Architecture Search Procedure of the Proposed Method}
\label{alg:propose}
    \begin{algorithmic}[1]
        \REQUIRE Dataset $\mathcal{D} = \{ \mathcal{D}_W, \mathcal{D}_\theta \}$
        \STATE Initialize $W$ and $\theta^{(1)}$, $\theta^{(2)}, \dots, \theta^{(N)}$
        \FOR{$t=1, \dots, T_W$}
            \STATE Sample mini-batch $\tilde{\mathcal{D}}_W$ from $\mathcal{D}_W$
            \STATE Sample $\lambda$ architectures from uniform distribution and update $W$ using \eqref{eq:omomi_gradient2}
        \ENDFOR
        \FOR{$t=1, \dots, T_\theta$}
            \STATE Sample mini-batch $\tilde{\mathcal{D}}_\theta$ from  $\mathcal{D}_\theta$
            \STATE Sample $\lambda$ architectures from $P_{\boldsymbol{\theta}}(M)$ and update $\theta^{(n)}$ for $n=1, \dots, N$ by \eqref{eq:is_theta_kousin}
        \ENDFOR
    \end{algorithmic}
\end{algorithm}

\subsection{Overall Algorithm}
The architecture search procedure followed by the proposed method is presented in Algorithm \ref{alg:propose}. The dataset $\D$ is divided into $\D_W$ and $\D_\theta$, and the resulting datasets are used to update the weights and distribution parameters, respectively. Although the method in \cite{Saito2019} jointly optimizes the weights $W$ and distribution parameters $\theta$, the proposed method separates the optimization of $W$ and $\theta$. That is, we first optimize $W$ under a uniform distribution and then optimize $\theta$ using the trained weights $W$. A separate (two-stage) optimization of the weight and architecture parameters has been conducted in recent NAS-related studies~\cite{chu2021scarlet,Guo2020eccv,sgnas}, and the approach has demonstrated promising performance.

In the optimization phase of $W$, $\lambda$ architecture parameters  $M_1,\dots,M_\lambda$ are sampled from a discrete uniform distribution, and the weight parameters of $W$ are updated using \eqref{eq:omomi_gradient2}. This update of $W$ is repeated $T_W$ times. Then, in the optimization phase of $\theta$, $\lambda$ architecture parameters $M_1,\dots,M_\lambda$ are sampled from the mixture distribution $P_{\boldsymbol{\theta}}(M)$, and the distribution parameters $\boldsymbol{\theta} = (\theta^{(1)}, \dots,\theta^{(N)})$ for different regularization coefficients $\epsilon_1, \dots, \epsilon_N$ are updated using \eqref{eq:is_theta_kousin}.
Following  the architecture search, we determine the final architectures by $M^{*}_{(n)}=\argmax_{M} P_{\theta^{(n)}}(M)$ for $n=1, \dots, N$ and obtain multiple architectures with different complexities. Then, we retrain the weights of the final architecture $M^{*}_{(n)}$ from scratch using the dataset $\D$.

\section{Experiment and Results}
This section evaluates the proposed method on image classification tasks. Our algorithms were run using NVIDIA Tesla V100 GPUs (32 GB memory).

\subsection{CIFAR-10}
\label{seq:cifar10_exp}

\subsubsection{Experimental Settings}
The CIFAR-10 \cite{Krizhevsky09learningmultiple} dataset contains 50,000 training and 10,000 test images, and each  image is labeled using one class out of 10. We adopt the cell-based CNN architecture search space used in \cite{AkimotoICML2019,Pham2018} and follow the experimental setting in \cite{AkimotoICML2019}. In the architecture search phase, we stack six normal and two reduction cells and set the number of channels in the first cell to 16. The architectures of the normal and reduction cells are searched by NAS algorithms. The training data $\D$ are divided into $\D_W$ and $\D_\theta$, which are then used to update the weights and distribution parameters, respectively. Both mini-batch sizes $|\tilde{\mathcal{D}}_W|$ and $|\tilde{\mathcal{D}}_\theta|$ are set to 64. We set the sample size of the architecture $\lambda$ to 2. The weights and distribution parameters are both updated for 200 epochs, respectively, i.e., $T_W = T_\theta = 200$. For updating the weights $W$, we use SGD with a momentum of 0.9 and set the weight decay to $3\times10^{-4}$. According to the cosine schedule~\cite{Loshchilov2017}, the learning rate gradually decreases from 0.025 to 0. For updating the distribution parameters $\theta$, we set the learning rate to $\eta_\theta = (\sum_{d=1}^{D} K_d)^{-1} = 1/180$ and the regularization coefficient $\epsilon_{n}$ to $\{0.0,0.1,0.3,0.5\}$. In the retraining phase, we set the number of normal cells to 10 and the number of channels in the first cell to 50. The other retraining settings are the same as \cite{AkimotoICML2019}. 

We compare the proposed method with two baseline one-shot NAS methods. The first method, presented in Algorithm~\ref{alg:comaprison1} and termed Method 1 (Simultaneous), is a straightforward extension of the method considered in \cite{Saito2019}. This algorithm simultaneously updates the weights and distribution parameters and performs the architecture search several times with different regularization coefficients $\epsilon$ to obtain multiple architectures. The second method, presented in Algorithm ~\ref{alg:comaprison2} and termed Method 2 (Separate), separates the weight optimization and architecture search, similar to the proposed method, but performs the architecture search several times with different $\epsilon$. The second method is advantageous compared to Method 1 (Simultaneous) because it performs the weight optimization only once; however, it is still inefficient compared to the proposed method because it requires multiple runs during the architecture search phase. The experiment uses the same number of epochs to optimize the weights and distribution parameters in a single search as in the proposed method, i.e., $T = T_W = T_\theta = 200$ in Algorithms~~\ref{alg:comaprison1} and \ref{alg:comaprison2}. Moreover, we perform the random search as the simplest baseline, which randomly samples architectures from the search space and retrains them. We sample architectures with weight parameters of 2.5M (million), 3.0M, 4.0M, and 5.0M. We reported the median values among three independent trials for all algorithms.

\begin{table}[t]
    \begin{minipage}[t]{0.49\linewidth}
        \begin{algorithm}[H]
            \caption{Method 1 (Simultaneous)}
            \begin{algorithmic}[1]
                \REQUIRE Dataset $\mathcal{D} = \{ \mathcal{D}_W, \mathcal{D}_\theta \}$
                \STATE Initialize $\theta^{(1)}$, $\theta^{(2)}, \dots, \theta^{(N)}$
                \FOR{$n=1, \dots, N$}
                    \STATE Initialize $W$
                    \FOR{$t=1, \dots, T$}
                        \STATE Sample mini-batch $\tilde{\mathcal{D}}_W$ from $\mathcal{D}_W$
                        \STATE Sample $\lambda$ architectures from $P_{\theta^{(n)}}(M)$ and update $W$ using \eqref{eq:omomi_gradient2}
                        \STATE Sample mini-batch $\tilde{\mathcal{D}}_\theta$ from  $\mathcal{D}_\theta$
                        \STATE Sample $\lambda$ architectures from $P_{\theta^{(n)}}(M)$ and update $\theta^{(n)}$ by \eqref{eq:theta_kousinsiki}
                    \ENDFOR
                \ENDFOR
            \end{algorithmic}
            \label{alg:comaprison1}
        \end{algorithm}
    \end{minipage}
    \hfill
    \begin{minipage}[t]{0.49\linewidth}
        \begin{algorithm}[H]
            \caption{Method 2 (Separate)}
            \begin{algorithmic}[1]
                \REQUIRE Dataset $\mathcal{D} = \{ \mathcal{D}_W, \mathcal{D}_\theta \}$
                \STATE Initialize $W$ and $\theta^{(1)}$, $\theta^{(2)}, \dots, \theta^{(N)}$
                \FOR{$t=1, \dots, T_{W}$}
                    \STATE Sample mini-batch $\tilde{\mathcal{D}}_W$ from $\mathcal{D}_W$
                    \STATE Sample $\lambda$ architectures from uniform distribution and update $W$ using \eqref{eq:omomi_gradient2}
                \ENDFOR
                \FOR{$n=1, \dots, N$}
                    \FOR{$t=1, \dots, T_{\theta}$}
                        \STATE Sample mini-batch $\tilde{\mathcal{D}}_\theta$ from  $\mathcal{D}_\theta$
                        \STATE Sample $\lambda$ architectures from $P_{\theta^{(n)}}(M)$ and update $\theta^{(n)}$ by \eqref{eq:theta_kousinsiki}
                    \ENDFOR
                \ENDFOR
            \end{algorithmic}
            \label{alg:comaprison2}
        \end{algorithm}
    \end{minipage}
\end{table}

\subsubsection{Results and Discussions}

\begin{figure}[t]
    \centering
    \begin{minipage}{0.6\linewidth}
        \centering
        \includegraphics[width=0.95\linewidth]{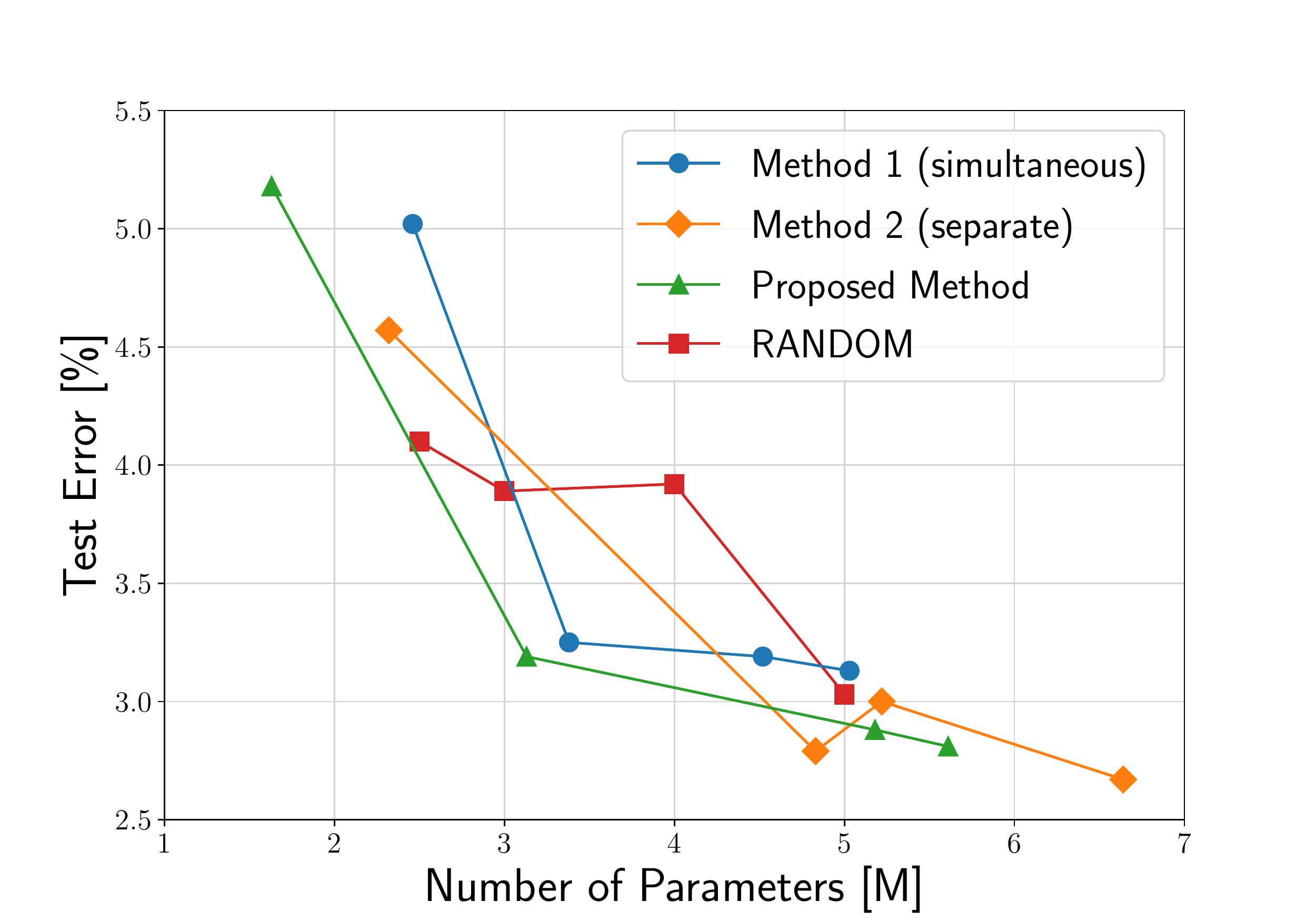}
        \caption{Relationship between the number of parameters and test error in CIFAR-10.}
        \label{fig:cifar10_result}
    \end{minipage}
    \hfill
    \def\@captype{table}
    \begin{minipage}{0.38\linewidth}
        \tblcaption{Search cost to obtain 4 architectures in CIFAR-10. The search cost excludes the architecture retraining cost.\\ }
        \label{table:cifar10_searchcost}
        \centering
        \begin{tabular}{cc}
        \toprule
        \multirow{2}{*}{Method} & Search Cost \\
         & $[\mathrm{GPU\:hours}]$ \\
        \hline
        Method 1 & \multirow{2}{*}{16.4} \\
        (Simultaneous) & \\
        \hline
        Method 2 & \multirow{2}{*}{5.2} \\
        (Separate) & \\
        \hline
        Proposed & \multirow{2}{*}{\textbf{3.4}} \\
        Method &   \\
        \bottomrule
        \end{tabular}
  \end{minipage}
\end{figure}

Figure~\ref{fig:cifar10_result} and Table~\ref{table:cifar10_searchcost} show the test error of the obtained architectures and search cost, respectively. The proposed method achieves better accuracies than those obtained by Method 1 (Simultaneous). In Method 1, the weight and distribution parameters are updated simultaneously. As the convergence speed of the operations in the cells differs, the distribution parameters converge to select the architecture that minimizes the loss early, resulting in a search failure. This difficulty associated with one-shot NAS during simultaneous optimization of weights and architectures has been pointed out in \cite{zhou2020NAS,chu2019fairdarts}. In comparison with Method 1, the proposed method selects well-performed architectures because all operations are equally selected and trained during the weight training stage. Table~\ref{table:cifar10_searchcost} shows that the proposed method obtains four architectures in approximately one-fifth of the search time required by Method 1. While Method 1 requires as many searches as the number of architectures to be obtained, the proposed method obtains multiple architectures in a single search.

Comparing the proposed method with Method 2 (Separate), the proposed method obtains multiple architectures with different parameters without causing a degradation of the prediction accuracy. Both methods require a single optimization of the weights, but the proposed method updates multiple distribution parameters using the architecture samples from the mixture distribution. Therefore, the proposed method does not need to repeat the architecture search. Consequently, the proposed method reduces the search cost compared to Method 2. Finally, the architectures obtained by the proposed method exhibit better prediction accuracies than those obtained via a random search, suggesting that the architecture search is effective.

\subsection{ImageNet}
\subsubsection{Experimental Settings}
ImageNet \cite{deng2009imagenet} is a large-scale image classification dataset consisting of 1,000 classes containing approximately 1.28 million training images and 50,000 validation images. We use the CNN architecture search space proposed in ProxylessNAS \cite{Cai2019} and evaluate the performance of the obtained architectures using the validation data. For the training data, we follow the pre-processing and data augmentation methods in \cite{sgnas}.

During the search phase, we update the distribution parameters $\theta$ with 50,000 randomly selected images from the training data $\D$ and update the weights $W$ with the remaining training data. We update the weights and distribution parameters for 60 epochs. We set the mini-batch sizes $|\tilde{\mathcal{D}}_W|$ and $|\tilde{\mathcal{D}}_\theta|$ to 350 and the number of samples $\lambda$ to 8. For updating the weights $W$, we use SGD with a momentum of 0.9 and set the weight decay to $5\times10^{-5}$. According to the cosine schedule \cite{Loshchilov2017}, the initial learning rate decreases from 0.068 to 0. For updating the distribution parameters, we set the learning rate and regularization coefficient to $\eta_\theta = (\sum_d^{D} K_d)^{-1} = 1/141$ and $\epsilon_{n} \in \{0.0,0.5,1.0\}$, respectively.

In the retraining phase, we update the weights for 350 epochs with a mini-batch size of 768. We use RMSProp and set the weight decay to $1 \times 10^{-5}$. In the first five epochs, the learning rate increases linearly from 0 to 0.192. Thereafter, the learning rate gradually decreases by multiplying 0.963 every three epochs. We use the label smoothing technique \cite{Szegedy2016} and introduce the squeeze and excitation module \cite{senet} into the MBConv operations. During inference, the model exponential moving average (EMA) is applied to calculate the prediction accuracy of the test data. These retraining settings are based on \cite{sgnas}.

\begin{table}[t]
  \caption{Result of ImageNet. $N$ represents the number of architectures to be searched.
  The values of the existing methods are referred from the literature.}
  \label{table:imagenet_result}
  \centering
  \begin{tabular}{lccc}
    \toprule
      Method &  Params  {\scriptsize $[\mathrm{M}]$} & Top-1 Accuracy {\scriptsize $[\%]$} & Search Cost {\scriptsize $[\mathrm{GPU\:hours}]$}
      \\
    \midrule
    MnasNet-A2~\cite{tan2019mnasnet} & 4.8 & 75.6 &  40,000$N$\\
    ProxylessNAS~\cite{Cai2019} & 4.4 & 75.3 &  200$N$ \\
    GreedyNAS-C~\cite{You_2020_CVPR} & 4.7 & 76.2 &   168+24$N$ \\
    SGNAS-C~\cite{sgnas} &  4.7 & 76.2 &  285 \\
    Proposed method ($\epsilon=1.0$)&  4.3 & 75.8 &   \textbf{164}\\
    \midrule
    MnasNet-A3~\cite{tan2019mnasnet} & 5.2 & 76.7 &   40,000$N$ \\
    GreedyNAS-B~\cite{You_2020_CVPR} & 5.2 & 76.8 &   168+24$N$ \\
    SGNAS-B~\cite{sgnas} &  5.5 & 76.8 &  285 \\
    Proposed method ($\epsilon=0.5$)&  5.4 & 76.8 &   \textbf{164}\\
    \midrule
    SCARLET-A~\cite{chu2021scarlet} & 6.7 & 76.9 &   240+48$N$ \\
    GreedyNAS-A~\cite{You_2020_CVPR} & 6.5 & 77.1 &   168+24$N$ \\
    SGNAS-A~\cite{sgnas} &  6.0 & 77.1 &  285 \\
    Proposed method ($\epsilon=0.0$)&  6.5 & 77.2 &   \textbf{164}\\
    \bottomrule
  \end{tabular}
\end{table}

\subsubsection{Results and Discussions}
Table~\ref{table:imagenet_result} shows the results of the proposed method and the existing NAS methods. The search cost indicates the cost to obtain $N$ optimized architectures. The proposed method ($\epsilon$ = 1.0) demonstrates a prediction accuracy of 75.8$\%$ with 4.4M parameters. This accuracy is worse than that of the existing methods; however, the number of parameters is lower than that in the existing methods. The prediction accuracies of the proposed method ($\epsilon$ = 0.5 and 0.0) are 76.8$\%$ and 77.2$\%$, respectively, indicating that these prediction accuracies are equal to or superior to those of existing methods.
The search cost of the proposed method is lower than that of existing NAS methods. MnasNet and ProxylessNAS require $N$ architecture searches to obtain $N$ architectures with varying complexities, similar to Method 1 described in Section \ref{seq:cifar10_exp}. GreedyNAS and SCARLET-NAS perform the architecture search multiple times after optimizing the supernet weights, similar to Method 2 described in Section \ref{seq:cifar10_exp}. Similar to the proposed method, SGNAS can obtain multiple structures in a single architecture search. However, SGNAS needs to train a DNN as the structure generator, which is more expensive than the proposed method. Our method results in a lower search cost compared with that in existing methods because it simultaneously updates multiple distributions by sharing the samples via importance sampling and realizes an efficient architecture search.

\section{Conclusion}
This paper has proposed a method for one-shot NAS that can efficiently find multiple architectures with different architecture complexities. We extended the method proposed in \cite{Saito2019} to be able to use categorical variables and have derived the natural gradient of the regularization term. Subsequently, we have proposed an efficient method to search multiple architectures via importance sampling. The experimental results produced using CIFAR-10 and ImageNet show that the proposed method obtains multiple well-performed architectures with different complexities by incurring less computational cost than the baseline methods.
Most NAS methods use fixed training hyperparameters, despite their impact on the performance. A possible future work could be developing a method for the joint optimization of both the architecture and training parameters, further improving the NAS performance.

\subsubsection{Acknowledgments}
This work was partially supported by NEDO (JPNP18002), JSPS KAKENHI Grant Number JP20H04240, and JST PRESTO Grant Number JPMJPR2133.
%
%
%
\bibliographystyle{splncs04}
\bibliography{bibliography}

\end{document}